# The AI Definition and a Program Which Satisfies this Definition

Dimiter Dobrev
Institute of Mathematics and Informatics
Bulgarian Academy of Sciences
*d@dobrev.com*

We will consider all policies of the agent and will prove that one of them is the best performing policy. While that policy is not computable, computable policies do exist in its proximity. We will define AI as a computable policy which is sufficiently proximal to the best performing policy. Before we can define the agent's best performing policy, we need a language for description of the world. We will also use this language to develop a program which satisfies the AI definition. The program will first understand the world by describing it in the selected language. The program will then use the description in order to predict the future and select the best possible move. While this program is extremely inefficient and practically unusable, it can be improved by refining both the language for description of the world and the algorithm used to predict the future. This can yield a program which is both efficient and consistent with the AI definition.

## 1. Introduction

Once, I was talking to a colleague and he told me: *'Although we may create AI someday, it will be a grossly inefficient program as we will need an infinitely fast computer to run it'*. My answer was: *'You just give me this inefficient program which is AI, and I will improve it so that it becomes a true AI which can run on a real-world computer'*.

Today, in this paper I will deliver the kind of program I asked my colleague to give me at that time. I will set out an inefficient program which satisfies the AI definition. I will go further and suggest some ideas and guidance on how this inefficient program can be improved to become a real program which runs in real time. My hope is that some readers of this paper will succeed to do this and deliver the AI we are looking for.

How inefficient is the program described here? In theory, there are only two types of programs – ones which halt and ones which run forever. In practice however, some programs will halt somewhere in the future, but they are so inefficient that we can consider them as programs which run forever. This is the case with the program described here — formally it halts, but its inefficiency makes it unusable (unless the computer is infinitely fast or the world is extremely simple).

**Tricks.** In this paper we will use some tricks which engineers will not be happy with. Mathematicians often prove the existence of non-constructive objects. These are well-defined objects which do exist, but cannot be displayed. In this paper we will define "the best policy" which is uncomputable and therefore non-constructive. Algorithms are descriptions of computable policies, however, our policy will not have its algorithm. In the eyes of engineers such a policy is meaningless because they only deal with constructive objects.

We will also introduce a computable policy which sufficiently approximates the best policy. Although it is a constructive object, engineers will not like it, either, because it will be computed by a program which requires an infinitely fast computer. Hence, from an engineer's perspective the policy will not be computable because a real computation is one which takes place in real time. Thus, we will define AI as a program the efficiency of which is unimportant, while in real life we only care about programs which can be used in real time (or in reasonable time).

Another trick is that we will use infinite objects which are much favored by mathematicians but do not sit well with engineers. Why do mathematicians prefer infinite objects although we do not see such objects in real life? Because things become a lot more interesting when we face infinity. For example, all finite functions are computable. If we need uncomputable functions, we must embrace infinity. While all finite functions can be described, the infinite functions are continuum many and only a countable part of them can be described. Infinity makes things more interesting as well as more simple. This is why we perceive the computer as a Turing machine (as an infinite function) although in reality a computer is a finite-state machine. Things become far more simple if we imagine that the computer has unlimited memory and computes infinite functions. Similarly, our understanding of AI will benefit a lot if we simply assume that its lifespan is unlimited.

**What is the definition of AI?** We will define AI as a policy. An agent who follows this policy will cope sufficiently well. This is true for any world, provided however that there are not any fatal errors in that world. If a fatal error is possible in a given world, the agent may not perform well in that particular world, but his average performance over all possible worlds will still be sufficiently good.

Which worlds we will consider as possible? The world's policies are continuum many. If we do not have any clues as to what the world should be, then we cannot have a clue about what the expected success of the agent should look like. We will assume that the world can be described and such description is as simple as possible (this assumption is known as *Occam's razor*). In other words, we will choose a language for description of worlds and will limit our efforts only to the worlds described by that language. The worlds whose description is simpler (shorter) will be preferred (will carry more weight).

This paper will consider several languages for description of the world. The first language will describe deterministic worlds. This language will describe the world by means of a computable function, which will take the state of the world and the action of the agent as input and return the new state of the world and the next observation as output. If we know the initial state of the world and agent's actions, this function will give us the life of the agent in that world.

The second language will describe non-deterministic worlds – again by a computable function, but with one additional argument. This argument will be randomness. In this case, we will need to know one more thing in order to obtain the agent's life in that world. We will need to know what that randomness has been.

We will define AI by these two languages and will make the assumption that these two definitions are identical. We will make even the assumption that the AI definition does not depend on our choice of language for description of worlds, and all languages produce the same definition of AI.

On the basis of these two languages we will make two programs which satisfy the AI definition. These two programs will calculate approximately the same policy, but their efficiency would be dramatically different. Therefore, the choice of language for description of the world will not affect the AI definition, but will have a strong impact on the efficiency of the AI obtained through the chosen language.

**Contributions**

This paper improves the AI definition initially provided by Hernández-Orallo et al. in 1998 (Orallo, 1998) and then substantially improved by Marcus Hutter in 2000 (Hutter, 2000). More precisely, this paper introduces two improvements:

**1. An AI definition which does not depend on the length of life.** Papers (Orallo 1998 and Hutter 2000) do provide an AI definition, however, the assumption there is that the length of life is limited by a constant and this constant is a parameter of the definition.

**2. An AI definition which does not depend on the language for description of the world.** The language in (Orallo 1998 and Hutter 2000) is fixed. Thus, these papers imply that there is only one possible way to describe the world.

## 2. Related work

### 2.1 General Intelligence

Let us first note that the meaning which we imply in *artificial intelligence* in this paper is *artificial general intelligence*. Other authors have discussed two types of AI which they describe as *narrow* and *general* (sometimes as *weak* and *strong*). I believe that a more appropriate pair of terms for the two types of AI is *fake* and *genuine AI*.

Let us illustrate this statement using the example of diamonds. Both intelligence and diamonds are classified in two categories – natural and artificial. Artificial diamonds are further divided in two subcategories – *genuine* (consisting of carbon) and fake (made of glass). Today, when we say artificial diamonds we mean ones made of carbon. Now let us image that we are living in the 19th century when nobody was yet able to make artificial diamonds from carbon. What people in the 19th century meant by artificial diamonds were diamonds made of glass – shiny pieces that look like diamonds but in fact are not. Today we call these glass pieces fake diamonds.

A genuine artificial diamond is every bit as good as a natural diamond. In terms of hardness and transparency these two diamonds are equal. However, they differ in price because an artificial diamond is much cheaper than a natural one although it may be superior in terms of size and purity.

The same applies to artificial intelligence. Artificial general intelligence is by all measures as good as natural intelligence, and can even be better in terms of speed, memory and "smartness". Certainly, the price of artificial intelligence will be much lower than that of natural intelligence. Today, in the 21st century, natural intelligence is even priceless because you cannot buy it.

Regarding narrow artificial intelligence, it looks like intelligence, but it is not. When we come to have artificial general intelligence one day, narrow AI programs will be called *fake artificial intelligence* or *intelligence-mimicking programs*.

Nowadays most papers dedicated to AI actually mean some narrow or fake AI. In this paper by AI we will mean general or genuine AI.

## 2.2 The Intuitive Definition

Now let us proceed with an overview of the papers dedicated to the definition of artificial intelligence. This definition is very important and actually drills down to the most important question about AI. Nonetheless, these papers a very few because most researchers never bother themselves with the question “What is AI?” – there are just a few researchers who do. The reason is that our colleagues simply do not believe in AI. If you do not believe in ghosts you do not ask yourself “What is actually a ghost?”. Recently I attended a lecture given by one of the leading experts in the area of AI (Solar-Lezama, 2023). He said “No matter how smart AI is, there will always be some human who is smarter than it”. Evidently, this colleague of ours does not believe in AI and cannot imagine that one day AI will be smarter than any human.

Although the papers dedicated to the AI definition are not so many, there are still some of them. Very good overviews of these papers can be seen in Wang 2019 and in the works of Hernández-Orallo (2012, 2014a, 2014b, 2014c, 2017). Here we will offer a shorter overview in which we will try to say things that have not been said in the mentioned overview papers.

The first intuitive (informal) definition of AI was provided by Alan Turing and is known as the Turing Test (Turing, 1950). That definition is perfect in its simplicity. Nonetheless, there is a significant problem with it. What the Turing Test defines is trained intellect (i.e. intelligence plus education). We would like have a definition of untrained intellect (i.e. pure intelligence without education). The first definition of pure intelligence was provided by Pei Wang in 1995 (Wang, 1995). It reads as follows:

*Intelligence is the capacity of an information-processing system to adapt to its environment while operating with insufficient knowledge and resources.*

This definition is very important because it is the first definition that separates intelligence from education. Later, in 2000, a simplified (refined) version of Pei Wang’s definition appeared. That version was published in Dobrev (2000). Today, it is the first result listed by *Google* on the topic of AI Definition. The first result returned by *Google* in response to a query for *Definition of Artificial Intelligence* is the paper of Dobrev (2005a), which is an improved version of Dobrev (2000). Here is the simplified version of Pei Wang’s definition:

*AI will be such a program which in an arbitrary world will cope not worse than a human.*

Which version of the definition of AI is better? Just because the simplified version comes out on top does not mean it is better than Pei Wang’s definition. Google simply prefers shorter papers and cleaner concepts.

What is the difference between the two versions of the definition? The first difference is that Pei Wang defines intelligence, while the simplified version defines artificial intelligence. This is not significant, because the real question is “What is intelligence?”. The fact that AI is a program is a direct corollary from Church thesis (Church, 1941) which says that any information system can be emulated by a computer program.

Here is another difference between the two versions of the definition: While Wang wants the intelligence to be able to cope in a concrete world (in its environment), according to the simplified version the intelligence must be able to cope in an arbitrary world. Why is this important? In the end of the day, for us it is important that AI is able to cope well in its own environment, because this is the important environment we are interested in. However, AI should not be dependent on the environment because we wish to be able to deploy it in various environments (worlds) such that each deployment is successful regardless of the environment. Although we can perfectly say that the real world is what matters to us, this world is not unique. The place and time of birth make a big difference. If either of these parameters were to change, we would find ourselves in a very different world. Obviously, Pei Wang was clearly mindful that there is not just one world, which is why he added to his definition the phrase *while operating with insufficient knowledge and resources*. I.e. Pei Wang wants AI to be able to cope in difficult circumstances as well, implying that if it succeeds when it is difficult it will also succeed when it is easy. Of course, things are difficult for those who are uneducated and poor. It would be much easier when one is equipped with knowledge and resources.

Another difference is that Pei Wang's definition does not say how well the AI should cope. Wang implies that AI will either cope or fail, but we know that some cope better than others. That is, how well AI can cope, and therefore its level of intelligence, is important. The simplified version of the definition says that AI should cope not worse than a human. Although benchmarking to a human makes the definition informal, it is still important because we should identify the level of intelligence which is sufficient for us to accept that a given program covers the necessary level of intelligence to be recognized as AI.

There is another difference between the two versions of the definition, and it is significant. In the simplified version, we assume that AI is any program whose input/output meets certain requirements. That is, we assume that AI is a class of programs without caring about efficiency (whether the program can run in reasonable time). The concept of reasonable time cannot be defined formally, because it is relative and depends on the power of the computer we will use. However, the concept of reasonable time, although informal, is intuitively clear. There are algorithms that theoretically work, but are so inefficient that they are practically unusable. For example, the algorithm for factoring large numbers into their prime factors is so inefficient that this is even used for data encryption.

Pei Wang's definition states that AI must be able to work with insufficient resources. This refers to both the resources of the environment (food, money) and the resources of the computer we use. That is, the resources of the environment and the resources of the AI itself (its memory and speed) are limited. That is, Pei Wang's definition recognizes as AI only those programs that can run in reasonable time.

### 2.3 One Discussion

A very serious discussion around Pei Wang's definition has been made in *Journal of Artificial General Intelligence, Volume 11 (2020): Issue 2 (February 2020), Special Issue "On Defining Artificial Intelligence" – Commentaries and Author's Response*.

Shane Legg noted in the discussion that it is not mandatory to have a definition of AI (Legg, 2020). He asserts that economists do not have an exact definition of what is economy, but that does not prevent them from developing their science. We cannot agree with this assertion. Economists deal with something which exists, while we are trying to create something which does not exists yet. Therefore, we are required to find the answer to the question "What is AI" because otherwise we would never know whether we have found the thing we are looking for.

Richard Sutton in (Sutton, 2020) draws our attention to John McCarthy's definition:

*Intelligence is the computational part of the ability to achieve goals in the world.*

It is legitimate to say that McCarthy's definition repeats Wang's definition, but expresses it in other words. We can accept that "*adapt to its environment*" is synonymous to "*achieve goals in the world.*" In any case we must be able to say when a given program copes better than another program. Whether we would prefer to call this coping *adaptation* or *achievement of goals* is not important.

Nevertheless, there is something in Sutton's reasoning which we definitely cannot agree with. Sutton puts an equality sign between the skill to solve a concrete problem and the skill to solve any problem. In his examples, thermostat and chess-playing program are such concrete problems. Programs which solve concrete problems are not intelligent. Intelligence is the ability to solve any problem. We spot the same issue also in John Laird (Laird, 2020). He asserts that Chinook, Deep Blue, and Watson are intelligent programs, but they are not intelligent because each of them solves a concrete problem rather than any problem.

Roger Schank says that computers cannot be intelligent (Schank, 2020). We fully agree with him. AI is a program. Even the most powerful and fastest computer will look stupid if we let it run a stupid program. Sutton also says that "*AI is now just about counting.*" Indeed, in our area today there is some hype about hyper-powerful computations, however, these computations already look smart and need just a little bit to become truly intelligent.

François Chollet says that the definition should measure the "degree of intelligence" (Chollet, 2020). We agree with this. As we said above, there should be different levels of intelligence.

Joscha Bach notes that in Wang's definition AI depends on the environment in which it is placed (Bach, 2020). Indeed, we also noted that AI should be able to cope in any environment.

Tomáš Mikolov and Roman Yampolskiy observe that we perceive AI as a separate being rather than something created by man which must serve man (Mikolov, 2020) and (Yampolskiy, 2020). On one side, we would agree with them, but on the other side we would say that not everything should be considered as be being a product created by man. We have heard environmental activists say that "We are eradicating many animal species, but the body of some of these animals may contain a priceless medicine which can cure many people from their diseases". To these environmental activists we will say that all living creatures have the right to live for their own sake and that they do not exist in order to satisfy some needs of ours. The same applies to AI. This is a notion which exists independently from man. Whether AI will be useful to us and whether it will work for us or we will work for it is a matter which depends on how we construct AI and manage to keep it under our control.

Alan Winfield draws our attention to the existence of various types of intelligence (Winfield, 2020). This is correct. When looking at humans, we have all observed that various people cope well with some tasks, and struggle with others. For example, there are very good mathematicians who are quite inapt in their social interactions. Winfield refers to *social intelligence*. Very important for this intelligence is that the model of the world includes more agents. I.e., the transition from a single-agent to a multi-agent model of the world is essential for social intelligence. In the present paper we will address this notion and will consider a language for description of worlds in which there are multiple agents. Peter Stone also says that there are different types of intelligence (Stone, 2020). He even goes further to insist that there should be different definitions for the different types of intelligence. Here we cannot agree with him. For example, there are various types of motor vehicles. Although there are dozens of types of trucks, racing cars, etc., this has not been an obstacle for having a common definition of a motor vehicle.

John Fox suggests that we should narrow down the set of possible worlds and focus on the area of medicine (Fox, 2020). On the upside he is right in saying that medicine is a fairly complicated area and if we manage to create a program that copes in this area it will probably cope in any area. On the downside, the area of medicine is so complex that focusing on it would increase, rather than decrease, the difficulty of the problem.

Raúl Rojas tells us that the AI definition is like the horizon and the closer we get to it the more distant it becomes (Rojas, 2020). In the past we used to admire many things as AI, but now we do not think so because these problems are already solved by a computer, and indeed a computer does that better than man. In fact, this is the case with the definition of narrow AI. As regards the definition of Artificial General Intelligence, it is fixed and does not run afar from us. Raúl Rojas tells us that natural and artificial intelligence will never converge. We fully agree with him. Artificial intelligence will catch up with natural intelligence in the few areas where AI is still lagging behind. The result will be a form of intelligence which overshadows natural intelligence across all areas. I.e. rather than convergence we will see strong divergence.

Gianluca Baldassarre and Giovanni Granato suggest that we copy the human brain (Baldassarre, 2020). Indeed, bionics is a central method in engineering disciplines, however, it should not be overestimated. For example, aerospace engineers study birds, but their airplanes do not look like birds and fly much faster and higher than birds. They may well use some common principles, but we have not seen a modern aircraft which flaps its wings.

Aaron Sloman raises several interesting philosophical questions (Sloman, 2020). For example, he asks about feelings: Will AI have feelings? Indeed, human intelligence is based on feelings. Furthermore, people derive their motivations from feelings. Humans do not have a clear definition of the objective of their existence. We may assume that the objective of humans is survival and reproduction, but it is not embedded in natural intelligence and people do not recognize that this is what they are here for. Instead of an embedded objective, humans have instincts that lead to feelings and these feelings drive them to desires which are conducive to survival and reproduction. For example, the fear of heights is an instinct that leads to a form of fear which is conducive to survival. Sexual desire and love come in the same vein.

In the process of constructing AI, is it a good idea to use feelings in order to define the AI objectives? When AI plays chess, it strives to achieve victory. We can say that achieving victory is a pleasure for AI. Should we also add fear, envy, love and other feelings? For the sake of truth, living with an overly emotional person is difficult. Maybe it makes sense during the construction process to avoid making AI overly emotional.

Peter Lindes draws our attention to the fact that the term AI is used in two meanings (Lindes, 2020). The first meaning is AI as a being and the second one is AI as a science. Here we mean only the first meaning of this term.

Peter Lindes raises another interesting question by adding another hurdle to AI. We want AI to cope well even with insufficient knowledge and resources, but Linden adds more limitations in terms of memory capacity and computing power. It appears that, on top of everything, we want AI to be stupid as well – which seems exaggerated. Actually, memory capacity and computing power limitations do not mean that AI must be stupid. AI is a program and that program can be executed on various computer configurations. It may be executed on computer which has a larger memory and runs faster. A smarter program however would run even on a more basic computer. Therefore, the addition of these limitations makes perfect sense.

Istvan Berkeley draws our attention to the fact that nowadays the phrase Artificial Intelligence is used for marketing purposes and every merchant assures us that his merchandise comes with embedded AI (Berkeley, 2020). According to Berkeley, there are many programs which are AI, but do not satisfy Wang's definition. Actually these programs are not AI at all and should not be branded as AI, even though merchants willingly brand them as such.

Marek Rosa notes that we cannot let AI live in an arbitrary world because that world would be excessively complicated and AI would not be able to cope in it (Rosa, 2020). The problems to be solved by AI should come in appropriate sequence starting with the most simple ones. Difficult problems should only come when the simple problems have been solved. People live in a world with teachers who put to them problems in the right sequence. Furthermore, the teacher helps by showing how problems can be solved. Wang's definition says nothing about the teacher, but it is assumed that a teacher is an additional amenity and AI should be ready to make use of this amenity when it is available.

Matthew Crosby and Henry Shevlin remind us that we do not live alone but in society (Crosby, 2020). They note that a genial composer will starve to death if there are no other agents to feed him. Indeed, when we explore an arbitrary world, we assume that it is a multi-agent world (in the general case). In this paper we present a language which describes multi-agent worlds where the main ability of AI is to communicate and deal with all other agents.

Kristinn Thórisson asserts that the basis of intelligence is the ability to conceive invisible things or figuring things out (Thorisson, 2020). In the present paper we build on the same idea. We are trying to find a language for description of the world. What that language will describe is the hidden state of the world. Thus, describing the hidden state of the world means that we should be able to conceive or imagine the world.

William Rapaport asks whether intelligence is computable (Rapaport, 2020). Actually this question draws the borderline between AI believers and disbelievers. We belong to the cohort of believers, hence we maintain that intelligence is computable.

These were our brief remarks to the authors who contributed to the discussion. We wish to commend the organizers of the discussion for bringing together a lot of prominent scholars in that area who have provided many meaningful and interesting insights on what the AI definition should be.

Detailed answers to the questions raised in the discussion are provided in Wang (2020).

## 2.4 Natural Intelligence

When we talk about natural intelligence we mean human intelligence. Certainly, animals also possess intelligence and in certain parameters they even surpass human intelligence. The long-term memory of elephants is better than that of humans. Experiments have shown that the short-term visual memory of monkeys is much better than that of humans.

Human intelligence is distinguished by reasoning. There two types of reasoning: logical, which is multi-step reasoning, and recognition – associative reasoning, which is single-step. When it comes to recognition, computers have already surpassed humans. Owing to neuronal networks computers already recognize faces and voices much better than us, humans. Logical reasoning is the last area in which we, humans, are still ahead of computers.

Are animals capable of logical (multi-step) reasoning? Indeed, my grandfather, who was a biologist, conducted already in his time an experiment in which he taught hens to count (Dobrev, 1993). This means that animals are capable of logical (multi-step) reasoning and this has been known since long ago.

## 2.5 Logical Reasoning

What does it take for computers to become capable of multi-step (logical) reasoning? There must be a hidden state, i.e. there is a need for transition from *full observability* to *partial observability*. In multi-step reasoning, what changes at each step is the internal state of the world. Could we change the *observation* instead of the internal state? Basically yes, but with *full observability* we see too much and will need to separate some part of the *observation* and keep changing it in the logical reasoning process. It would be more natural to present the separated part of the observation as a hidden state of the world.

Logical reasoning requires "understanding". We must be able to understand "what is going on". This means that we must describe the hidden state of the world. For this purpose, we need some language for description of worlds. We can picture the hidden states of the world as elements of some countable set, as natural numbers or as words over some alphabet. The meaning of these words would give us the language for description of worlds.

Today the performance of chatbots such as *ChatGPT* (OpenAI, 2022) is amazing. Nevertheless, when we talk to them we get the feeling that they lack understanding. We are left with the unpleasant impression that we are talking to a parrot. Certainly, a chat with *ChatGPT* is incomparably more elaborate than talking to a parrot, but there is still room for improvement.

Moreover, in these chabots there is a degree of deception. For example, as per Yahav (2023), *ChatGPT* consists of two parts – a neural network and algorithms written by programmers. A neuronal network is incapable of multi-step reasoning, but *ChatGPT* misleads us to believe that it does multi-step reasoning owing to the added algorithms written by programmers. For example, the addition of two numbers takes multi-step reasoning and that operation is executed by the added algorithms. Why is this a deception? Because *ChatGPT* should be using only neural networks, or, if it does use additional programs, it should be able to create these programs itself rather than rely on the help of programmers. The issue here is not that the chatbot resorts to programmers. The issue is that each problem requires a separate patch and that it is not possible to write all patches that cover all problems.

A humanoid robot by the name *Sophia* was presented in 2015 (Retto, 2017). That robot also involved some deception. On one side, *Sophia* was misleading by its outer appearance, and on other side it had a remote control function. Although *Sophia* willingly talked to journalists, it was not clear at which moment it talked from its embedded AI and at which moment it relied on a human operator.

All AI definitions known to us consider AI as device with a memory (i.e. with an internal state), while the known implementations are based on neuronal networks and assume that AI does not need any memory (*full observability*). In other words, there is incoherence between definitions and implementations.

With regard to the internal state of AI we should note that what matters is the internal state of the world, while the internal state of AI only reflects the state of the world. Thus, the internal state of AI is actually AI's "perception" of the internal state of the world. Each change of the internal state of AI must be induced by the world. For example, if our AI "gets angry", that would be a change of its internal state, however, that change should be induced by the world. Our AI should not get angry without a reason. We wish to create AI which does not change its internal state frivolously, but only in response to information received from the world. More precisely, a new piece of information may not necessarily come directly from the world, but with a delay after a period of reflection.

## 2.6 The Formal Definition

The first formal definitions of AI were published in Hernández-Orallo (1998) and Hutter (2000). The definition in Orallo (1998) has many imperfections which were noted in Dobrev (2019b). Given these imperfections, we can assume that the first formal AI definition was provided by Marcus Hutter.

We only have one minor remark to Marcus Hutter's definition. Hutter defines AI as the best policy (he called it AIXI or AIξ). This is not good at least because AIξ is an uncomputable policy. It would be more appropriate to say that AI is a computable policy which is "near" the best one. We may even have to include an efficiency requirement because a program which is excessively inefficient is actually futile.

Hutter did propose a computable policy (AIXI*tl*) in Hutter (2007). This is a concrete algorithm which cannot be a definition of AI, either. Even if the AIXI*tl* algorithm were recognized as AI, it would not be the only algorithm which satisfies the AI definition. Any other algorithm which calculates the same policy would be AI as well, especially if it works more efficiently (faster) than AIXI*tl*. Moreover, the policy of AI need not necessarily be exactly the same as the policy of AIXI*tl*. It is enough for the policy to be sufficiently good.

While this minor remark applies to Marcus Hutter's definition, it does not apply to Dobrev (2005b and 2019b) because in those papers AI is defined as an arbitrary program the IQ of which exceeds a specified level.

The present paper contributes to the AI definition by introducing two improvements which apply to all formal AI definitions known to us to date.

## 2.7 The First Improvement

The first improvement relates to the length of life. Hernández-Orallo (1998) and Hutter (2000) assume that the length of life is limited. The same assumption was made in Dobrev (2005b and 2019b). However, many considerations suggest that it is desirable to avoid this assumption. Indeed, the lifespan of natural intelligence is limited, but this has nothing to do with intelligence itself. The lifespan of AI also may be limited, because eventually we will decide to shut it down. However, AI does not know when we are going to do this and should function steadily until the very last moment without bothering about the time at which shutdown will occur. Even if we assume that the length of life is limited by some constant $m$, this constant would be so big that we should better equate it to infinity.

If we assume that the length of life is limited, then AI would be a finite function. Why is it important to make a transition from finite to infinite functions? Because, as we already said, things become a lot more interesting when we face infinity.

Obviously, Hernández-Orallo and Marcus Hutter share our wish to avoid limiting the length of life, because both Hutter (2006) and Hernández-Orallo (2011) offer an improved version of the definition in which the upper bound is removed. This has been achieved by introducing a discount factor $\gamma$.

The discount factor $\gamma$ determines the notion of *greed*. This notion tells us whether our AI will aim for a quick win or would rather pursue success over a longer time frame. When $\gamma$ tends to 0 greed goes up and when $\gamma$ tends to 1 greed goes down.

It can be said that when a discount factor is used, the entire life is used for the calculation of successfulness, but this is not quite true. In practice, there comes a certain moment after which the impact of life on the success score becomes negligibly low.

This is illustrated by the following formula:

$$\forall \varepsilon > 0\ \forall \gamma\ \exists m \left( \left| 1 - \frac{Success(L_m)}{Success(L)} \right| < \varepsilon \right)$$

For each $\varepsilon > 0$ and for each discount factor $\gamma$ there exists some moment $m(\gamma)$ such that the part of life until moment $m(\gamma)$ determines main part of success, namely ($1-\varepsilon$), while the remaining part of success ($\varepsilon$) is determined by the infinite part of life which remains after moment $m(\gamma)$.

In this paper we have chosen another approach which uses, in a very substantial manner, the entire length of life. The best performing policy in our approach uses the limit to which the average score tends, and always selects an action which has the maximal limit. Thus, the best performing policy never makes fatal errors.

**Note:** The fact that we have selected a policy which does not make fatal errors does not mean that if we follow that policy we will walk the path which has the best possible average success. It means something else. Such a path will be available after each step, however, it is far from certain that in the end of the day we would have followed exactly that path. As an example I will provide a program which plays chess. My students and I wrote this program as a practical exercise. It calculated the next three moves and in this way it selected the best action. When the program sighted victory, it selected this action regardless of whether the victory would come after one, two or three moves. So the behaviour of our program became weird. Whenever the program saw a way to victory, rather than mating the opponent outright it kept playing cat and mouse with it. The program was always three moves away from victory, but it did not hurry to finish off the game. That weird effect disappeared as soon as we added some greed and made a victory that comes in one move more valuable than a victory that comes in two moves.

So, if we have two actions, and none of them leads to a fatal error, which one should be preferred as the best performing policy? In this paper we have decided that the choice will be based on maximum greed (say, based on an infinitely small discount factor $\gamma$). Another approach would be to use a fixed greed value ($0 < \gamma < 1$). We are not fond of this approach, either, because even when $\gamma$ is very close to 1, our AI would still be too greedy since it will remain too focused on how quickly success comes by.

Another deficiency of the greed-based approach is that AI will tend to needlessly prolong the actions whenever it expects to receive a negative reward. We humans often choose this approach – when we anticipate something bad to happen, we aim to push it away in time as much as we can. Nonetheless, in some cases we prefer not to procrastinate things. For example, when we realize that we are going to lose a chess game we would surrender rather than keep playing to the end.

Here is an idea how to define AI which is not greedy and at the same time does not beat around the bush. Let us say that if two paths lead to one and the same state, we will prefer the path that yields more success (it is important that we compare actual rather than average success because the length of the two paths may be different). If the two paths yield the same success, we will prefer the shorter path.

Thus, when AI realizes it is going to lose a chess game, it will surrender because there will be two possible paths that lead to the same state and the same success. In this case the success will be “a single loss”.

## 2.8 Additional Parameters

Greed is one of the additional parameters of AI. We have other additional parameters such as courage and curiosity. These parameters do not determine straightforwardly whether the success will increase or decrease. There are worlds in which being more greedy is better, while in other worlds greed is a disadvantage.

In humans, the values of these parameters are not the same across the board. There are situations in which courageous people survive as well as situations in which the more cautious ones win. If all people were the same, they would be at a risk of extinction because in a given situation they would all behave in the same way. Owing to the fact that people are different, they act in a different way and this is how part of the population always survives.

There are also basic parameters, such as memory and intelligence, which straightforwardly increase the successfulness of AI in an arbitrary world. We might design a special world which penalizes those who remember more or are more intelligent, but in most worlds memory and intelligence make a positive difference.

This is the reason why it would be better to take out the additional parameters from the definition. This would give us the freedom to choose the kind of AI we want to have – more courageous or more cautious. As regards the basic parameters, we will assume that their values are maximal and are only limited by the memory and the speed of the computer on which we will launch our AI.

## 2.9 The Second Improvement

The first improvement of the definition is not very significant. Far more important is the second improvement, namely that one of the most important parameters of the AI definition is the language for description of the world.

Admittedly, Marcus Hutter noted in (Hutter, 2007) that the universal Turing machine is a parameter of the definition:

*It (slightly) depends on the choice of the universal Turing machine.*

Hutter however suggests that the world is described by a computable function and puts an equality sign between programming languages and languages for description of worlds. In fact, the possible descriptions of the world are diverse and are not limited just to a description of a computable function.

In this paper we will consider various descriptions of worlds. First, we will look at the most standard presentation of the world as a deterministic computable function. Subsequently we will add randomness, then we will add some agents and eventually will end up with most diverse languages for description of worlds.

## 2.10 An Alternative Opinion

In a recent open letter Elon Musk (Musk, 2023) urged us to slow down and suspend AI research for six months. Perhaps not all research but in any case stop those experiments that may lead to a technogenic disaster. Basically Musk is right, but once the ghost is let out of the flask it is very hard to squeeze it back in. I agree that we should be very cautious with experiments, especially when we do not quite know what exactly their results would be. Most importantly, however, we should first ask ourselves what is actually AI and how are we going to live with it from now on.

## 2.11 What Is Happening Right Now?

We are on the verge of the emergence of Artificial General Intelligence (AGI). That discovery will transform our life and will make it very easy as well as totally meaningless. Our life will change a lot and we are not at all sure whether it will be a good change. (We will not provide a definition of what is good!)

Right now mankind is experiencing a spike of machine intelligence and people really got scared. The word is about the recent emergence of *ChatGPT*. Although this computer program is amazingly intelligent, it is not AI yet. Nevertheless, *ChatGPT* is just one very small step away from becoming *artificial general intelligence*. That missing step is the description of the hidden state. The present paper will describe exactly that very final step.

You might ask the following question: “If this is the final step remaining on the journey to AGI, why do you rush to make it? Why don’t you wait a little?”

The truth is that the emergence of AGI is inevitable. If we stop here and do not take part in its creation, our colleagues will not stop and will create it.

Some say that if the emergence of AI is inevitable then nothing can be done. In fact, there is something of great importance that will be done. We will be the generation which will select the rules for AI and will thereby shape the life of people for many years ahead or perhaps forever.

Once it is created, AGI will operate to the rules set by its creators. Importantly, these rules will be unchangeable because there will be a single AGI who will govern us and will not let us create another AGI.

## 3. Terms of the problem

Let the agent have $n$ possible actions and $m$ possible observations. Let $\Sigma$ and $\Omega$ be the sets of actions and respectively observations. In the observations set there will be two special observations. These will be the observations *good* and *bad*, and they will provide rewards 1 and -1. All other observations in $\Omega$ will provide reward 0.

We will add another special observation – *finish*. The agent will never see that observation ($finish \notin \Omega$), but we will need it when we come to define the model of the world. The model will predict *finish* when it breaks down and becomes unable to predict anything more. For us the *finish* observation will not be the end of life, but rather a leap in the unknown. We expect our AI to avoid such leaps in the unknown and for this reason the reward given by the *finish* observation will be -1.

We will begin with the tree of all possibilities. It is something like the infinite complete binary tree, however, its branches will not be just two, but $n$ or $m$. Another difference is that our tree will also have leafs.

**Definition 1:** The tree of all possibilities is an infinite tree. All vertices which sit at an even-number depth level and are not leafs will be referred to as action vertices and those at odd-number depth levels will be observation vertices. From each action vertex there will depart $n$ arrows which correspond to the $n$ possible actions of the agent. From each observation vertex there will depart $m+1$ arrows which correspond to the $m$ possible observations of the agent and the observation *finish*. The arrow which corresponds to *finish* will lead to a leaf. All other arrows lead to vertices which are not leafs.

**Definition 2:** In our terms the world will be a 3-tuple $<S, s_0, f>$, where:
1. $S$ is a finite or countable set of internal states of the world;
2. $s_0 \in S$ is the initial state of the world; and
3. $f: S \times \Sigma \rightarrow \Omega \times S$ is a function which takes a state and an action as input and returns an observation and a new state of the world.

The $f$ function cannot return observation *finish* (it is predicted only when $f$ is not defined and there is not any next state of the world). What kind of function is $f$ – computable, deterministic or total? The answer to each of these three questions can be *Yes*, but it can also be *No*.

**Definition 3:** A deterministic policy of the agent is a function which assigns a certain action to each action vertex.

**Definition 4:** A non-deterministic policy of the agent is a function which assigns one or more possible actions to each action vertex.

When the policy assigns all possible actions at a certain vertex (moment) we will say that at that moment the policy does not know what to do. We will not make a distinction between an agent and the policy of that agent. A union of two policies will be the policy which we get when choose one of these two policies and execute it without changing that policy. Allowing a change of the chosen policy will lead to something else.

**Definition 5:** Life in our terms will be a path in the tree of all possibilities which starts from the root.

Each life can be presented by a sequence of actions and observations:

$$a_1, o_1, \ \dots \ , a_t, o_t, \ \dots$$

We will not make a distinction between a finite life and a vertex in the tree of all possibilities because there is a one-to-one correspondence between these two things.

**Definition 6:** The length of life will be $t$ (the number of observations). Therefore, the length of life will be equal to the length of the path divided by two.

**Definition 7:** A completed life is one which cannot be extended. In other words, it will be an infinite life or a life ending with the observation *finish*.

When we let an agent in a certain world, the result will be a completed life. If the agent is non-deterministic then the result will not be unique. The same applies when the world is non-deterministic.

## 4. The grade

Our aim is to define the agent's best performing policy. For this purpose we need to assign some grade to each life. This grading will give us a linear order by which we will be able to determine the better life in any pair of lives.

Let us first determine how to measure the success of each life $L$. For a finite life, we will count the number of times we have had the observation *good*, and will designate this number with $L_{good}(L)$. Similar designations will be assigned to the observations *bad* and *finish*. Thus, the success of a finite life will be:

$$Success(L) = \frac{L_{good}(L) - L_{bad}(L) - L_{finish}(L)}{|L|}$$

Let us put $L_i$ for the beginning of life $L$ with a length of $i$. The *Success(L)* for infinite life $L$ will be defined as the limit of *Success*($L_i$) when $i$ tends to infinity. If this sequence is not convergent, we will take the arithmetic mean between the limit inferior and limit superior.

$$Success(L) = \frac{1}{2} . \left( \liminf_{i \to \infty} \big( Success(L_i) \big) + \limsup_{i \to \infty} \big( Success(L_i) \big) \right)$$

By doing this we have related each life to a number which belongs to the interval *[-1, 1]* and represents the success of this life. Why not use the success of life for the grade we are trying to find? This is not a good idea because if a world is free from fatal errors then the best performing policy will not bother about the kind of moves it makes. There would be one and only one maximum success and that success would always be achievable regardless of the number of errors made in the beginning. If there are two options which yield the same success in some indefinite time, we would like the best performing policy to choose the option that will yield success faster than the other one. Accordingly, we will define the grade of a completed life as follows:

**Definition 8:** The grade of infinite life $L$ will be a sequence which starts with the success of that life and continues with the rewards obtained at step $i$:

$$Success(L),\ reward(o_1),\ reward(o_2),\ reward(o_3),\ \ldots$$

**Definition 9:** The grade of finite and completed life $L$ will be the same sequence, but in this sequence for $i>t$ the members $reward(o_i)$ will be replaced with *Success(L)*:

$$Success(L),\ reward(o_1),\ \ldots\ ,\ reward(o_t),\ Success(L),\ Success(L),\ \ldots$$

(In other words, the observations that come after the end of that finite life will receive some expectation for a reward and that expectation will be equal to the success of that finite life.)

In order to compare two grades, we will take the first difference. This means that the first objective of the best performing policy will be the success of entire life, but its second objective will be to achieve a better reward as quickly as possible.

## 5. The expected grade

**Definition 10:** For each deterministic policy $P$ we will determine *grade(P)*: the grade we expect for the life if policy $P$ is executed.

We will determine the expected grade at each vertex $v$ assuming that we have somehow reached $v$ and will from that moment on execute policy $P$. The expected grade of $P$ will be the one which we have related to the root.

We will provide a rough description of how we relate vertices to expected grades. Then we will provide a detailed description of the special case in which we look for the best grade, i.e. the expected grade of the best performing policy.

Rough description:

1. Let $v$ be an action vertex.
Then the grade of $v$ will be the grade of its direct successor which corresponds to action $P(v)$.

2. Let $v$ be an observation vertex.
2.1. Let there be one possible world which is a model of $v$.
If we execute $P$ in this world we will get one possible life. Then the grade of $v$ will be the grade of that life.
2.2. Let there be many possible worlds.
Then each world will give us one possible life and the grade $v$ will be the mean value of the grades of the possible lives.

The next section provides a detailed description of the best performing policy. The main difference is that when $v$ is an action vertex, the best performing policy always chooses the highest expected grade among the expected grades of all direct successors.

## 6. The best performing policy

As mentioned above, we should have some clue about what the world looks like before can have some expectation about the success of the agent. We will assume that the world can be described by some language for description of worlds.

Let us take the standard language for description of worlds. In this language the world is described by a computable function (this is the case in Orallo, 1998 and Hutter, 2000). We will describe the computable function $f$ by using a Turing machine. We will describe the initial state of the world as a finite word over the machine alphabet. What we get is a computable and deterministic world which in the general case is not a total one.

**Definition 11:** A world of complexity $k$ will be a world in which:
1. The $f$ function is described by a Turing machine with $k$ states.
2. The alphabet of that machine contains $k+1$ symbols ($\lambda_0, \dots, \lambda_k$).
3. The initial state of the world is a word made of not more than $k$ letters. The alphabet is $\{\lambda_1, \dots, \lambda_k\}$, i.e. the alphabet of the machine without the blank symbol $\lambda_0$.

Here we use the same $k$ for three different things as we do not need to have different constants.

We will identify the best performing policy for the worlds of complexity $k$ (importantly, these worlds are finitely many). For this purpose we will assign to each observation vertex its best grade (or the expected grade if the best performing policy is executed from that vertex onwards).

Let us have life $a_1, o_1, \dots , a_t, o_t, a_{t+1}$.
Let this life run through the vertices $v_0, w_1, v_1, \dots , w_t, v_t, w_{t+1}$,
where $v_0$ is the root, $v_i$ are the action vertices and $w_i$ are the observation vertices.

Now we have to find out how many models of complexity $k$ are there for vertex $v_t$.

**Definition 12:** A deterministic world is a model of $v_t$ when in that world the agent would arrive at $v_t$ if he executes the corresponding actions ($a_1, \dots , a_t$). The models of each action vertex are identical with the models of its direct successors.

**Definition 13:** The best performing policy for the worlds of complexity $k$ will be the one which always chooses the best grade (among the best grades of the direct successors).

**Definition 14:** The best grade of vertex $w_{t+1}$ (for worlds of complexity $k$) is determined as follows:

**Case 1.** Vertices $v_t$ and $w_{t+1}$ do not have any model of complexity $k$.
In this case the best grade for $w_{t+1}$ will be *undef*. At this vertex the policy will not know what to do (across the entire subtree of $v_t$) because the best grade for all successor vertices will be *undef*.

If we do not want to introduce an *undef* grade, we can use the lowest possible grade – the sequence of countably many -1s. The maximal grade will be chosen among the vertices which are different from *undef*. Replacing *undef* with the lowest possible grade will give us the same result.

**Case 2.** Vertices $v_t$ and $w_{t+1}$ have one model of complexity $k$.
Let this model be $D$. In this case there are continuum many paths through $w_{t+1}$ such that $D$ is model of all those paths. From these paths (completed lives) we will select the set of the best paths. The grade we are looking for is the grade of these best paths. Each of these paths is related to a deterministic policy of the agent. We will call them the best performing policies which pass through vertex $w_{t+1}$.

This is the procedure by which we will construct the set of best deterministic policies: Let $P_0$ be the set of all policies which lead to $w_{t+1}$. We take the success of each of these policies in the world D. We create the subset $P_1$ of the policies which achieve the maximum success. Then we reduce $P_1$ by selecting only the policies which achieve the maximum for $reward(o_{t+2})$ and obtain subset $P_2$. Then we repeat the procedure for each $i>2$. In this way we obtain the set of the best deterministic policies $P$. (The best ones of those which pass through vertex $w_{t+1}$ as well as the best ones for the paths which pass through vertex $w_{t+1}$. As regards the other paths, it does not matter how the policy behaves there.)

$$P = \bigcap_{i=0}^{\infty} P_i$$

We can think of $P$ as one non-deterministic policy. Let us take some $p \in P$. This will give us the best grade:

$$Success(p), reward(o_{t+1}), reward(o_{p,t+2}), reward(o_{p,t+3}), \ldots$$

Here we drop out the members $reward(o_i)$ at $i \leq t$ because they are uniquely defined by $v_t$. The next member depends on $w_{t+1}$ and $D$, but does not depend on $p$. The remaining members depend on $p$.

Another way to express the above formula is:

$$\max_{p \in P_0} Success(p), reward(o_{t+1}), \max_{p \in P_1} reward(o_{p,t+2}), \max_{p \in P_2} reward(o_{p,t+3}), \ldots$$

**Case 3.** Vertices $v_t$ and $w_{t+1}$ have a finite number of models of complexity $k$.
Let the set of these models be $M$. Again, there are continuum many paths through $w_{t+1}$ such that each of these paths has a model in $M$. These paths again form a tree, but while in case 2 the branches occurred only due to a different policy of the agent, in this case some branches may occur due to a different model of the world. Again, we have continuum many deterministic policies, but now they will correspond to subtrees (not to paths) because there can be branches because of the model. Again we will try to find the set of best performing deterministic policies and the target grade will be mean grade of those policies (the mean grade in *M*).

We will again construct the set of policies $P_i$. Here $P_1$ will be the set of policies for which the mean success reaches its maximum. Accordingly, $P_2$ will be the set of policies for which the mean $reward(o_{t+2})$ reaches its maximum and so on. This is how the resultant grade will look like:

$$\max_{p \in P_0} \sum_{m \in M} q_m . Success(m,p), \sum_{m \in M} q_m . reward(o_{m,t+1}), \max_{p \in P_1} \sum_{m \in M} q_m . reward(o_{m,p,t+2}), \ldots$$

If we take some $p \in P$, the resultant grade will look like this:

$$\sum_{m \in M} q_m . Success(m,p), \sum_{m \in M} q_m . reward(o_{m,t+1}), \sum_{m \in M} q_m . reward(o_{m,p,t+2}), \ldots$$

Here $q_i$ are the weights of the worlds which have been normalized in order to become probabilities. In this case we assume that the worlds have equal weights, i.e.:

$$q_i = \frac{1}{|M|}$$

■

What we have described so far looks like an algorithm, however, rather than an algorithm, it is a definition because it contains uncomputable steps. The so described policy is well defined, even though it is uncomputable. Now, from the best grade for complexity *k*, how can we obtain the best grade for any complexity?

**Definition 15:** The best grade at vertex *v* will be the limit of the best grades at vertex *v* for the worlds of complexity *k* when *k* tends to infinity.

How shall we define the limit of a sequence of grades? The number at position *i* will be the limit of the numbers at position *i*. When the sequence is divergent, we will take the arithmetic mean between the limit inferior and limit superior.

**Definition 16:** The best performing policy will be the one which always chooses an action which leads to the highest grade among the best grades of the direct successors.

What makes the best performing policy better than the best performing policy for worlds of complexity *k*? The first policy knows what to do at every vertex, while the latter does not have a clue at the majority of vertices because they do not have any model of complexity *k*. The first policy can offer a better solution than the latter policy even for the vertices at which the latter policy knows what to do because the first policy also considers models of complexity higher than *k*. Although at a first glance we do not use Occam's razor (because all models have equal weights), in earnest we do use Occam's razor because the simpler worlds are calculated by a greater number of Turing machines, meaning that they have a greater weight.

## 7. The AI definition

**Definition 17:** AI will be a computable policy which is sufficiently proximal to the best performing policy.

At this point we must explain what makes a policy proximal to another policy and how proximal is proximal enough. We will say that two policies are proximal when the expected grades of these two policies are proximal.

**Definition 18:** Let $A$ and $B$ be two policies and $\{a_n\}$ and $\{b_n\}$ are their expected grades. Then the difference between $A$ and $B$ will be $\{\varepsilon_n\}$, where:

$$\varepsilon_n = \sum_{i=0}^{n} \gamma^i (a_i - b_i) = \varepsilon_{n-1} + \gamma^n (a_n - b_n)$$

Here $\gamma$ is a discount factor. Let $\gamma=0.5$. We have included a discount factor because we want the two policies to be proximal when they behave in the same way for a long time. The later the difference occurs in time, the less impact it will have.

When $n$ goes up, $|\varepsilon_n|$ may go up or down. We have made the definition in this way because we want the difference to be small when the expected grade of policy $A$ hovers around the expected grade of policy $B$. I.e., if for $n-1$ the higher expected grade is that of $A$ and for $n$ the higher expected grade is that of $B$, then in $\varepsilon_n$ the increase will offset the decrease and vice versa.

**Definition 19:** We will say that $|A-B|<\varepsilon$ if $\forall n\ |\varepsilon_n|<\varepsilon$.

## 8. A program which satisfies the definition

We will describe an algorithm which represents a computable policy. Each action vertex relates to an uncompleted life and the algorithm will give us some action by which this life can continue. This algorithm will be composed of two steps:

**1. The algorithm will answer the question 'What is going on?'** It will answer this question by finding the first $k$ for which the uncompleted life has a model. The algorithm will also find the set $M$ (the set of all models of the uncompleted life, the complexity of which is $k$). Unfortunately, this is uncomputable. To make it computable we will try to find efficient models with complexity $k$.

**Definition 20:** An efficient model with complexity $k$ will be a world of complexity $k$ (definition 11), where the Turing machine uses not more than $1000.k$ steps in order to make one step of the life (i.e. to calculate the next observation and the next internal state of the world). When the machine makes more than $1000.k$ steps, the model will return the observation *finish*.

The number *1000* is some parameter of the algorithm, but we assume this parameter is not very important. If a vertex has a model with complexity $k$, but does not have an efficient model with complexity $k$, then $\exists n\ (n>k)$ such that the vertex has an efficient model with complexity $n$.

**2. The algorithm will answer the question 'What should I do?'.** For this purpose we will run $h$ steps in the future over all models in $M$ and over all possible actions of the agent. In other words, we will walk over one finite subtree and will calculate *best* for each vertex of the subtree (this is the best expected grade up to a leaf). Then we will choose an action which leads to the maximum by *best* (this is the best partial policy).

**Definition 21:** A partial subtree of vertex $v_t$ over $M$ with depth $h$ will be the subtree of $v_t$ composed of the vertices which i) have a depth not more than $2h$ and ii) have a model in $M$.

**Definition 22:** The grade up to a leaf of vertex $v_{t+i}$ to the leaf $v_{t+j}$ will be:

Case 1. If $j=h$, this will be the sequence:

$$Success(v_{t+j}),\ reward(o_{t+i+1}),\ \dots\ ,\ reward(o_{t+j})$$

Case 2. If $j<h$, then the sequence in case 1 will be extended by $h-j$ times $Success(v_{t+j})$. The purpose of this extension is to ensure that the length of the grade up to a leaf will always be $h-i+1$.

**Definition 23:** The best expected grade up to a leaf (this is *best*):

1. Let $v_{t+i}$ be an action vertex.

1.1. If $v_{t+i}$ is a leaf, then $best(v_{t+i})$ will be the grade up to a leaf of $v_{t+i}$ to the leaf $v_{t+i}$.

1.2. If $v_{t+i}$ is not a leaf then:

$$best(v_{t+i}) = \max_{a \in \Sigma} best(w_a)$$

By $w_a$ here we designate the direct successor of $v_{t+i}$ resulting from action $a$. The same applies accordingly to $v_o$ below.

2. Let $w_{t+i}$ be an observation vertex. Then:

$$best(w_{t+i}) = \sum_{o \in \Omega'} p_o . (reward(o) \text{ insert_at_1_in } best(v_o))$$

Thus, we take the *best* of the direct successor $v_o$ and extend it by one by inserting $reward(o)$ at position 1. Here $\Omega' = \Omega \cup \{finish\}$ and $p_o$ is the probability of the next observation being $o$. Let $M(v)$ be the set of the models of $v$. Then:

$$p_o = \frac{\left(\sum_{m \in M(v_o)} q_m\right)}{\left(\sum_{m \in M(w_{t+i})} q_m\right)} = \frac{|M(v_o)|}{|M(w_{t+i})|}$$

In this formula $q_m$ are the weights of the models. The last equality is based on the assumption that all models have equal weights. If $M(v_o)=\emptyset$ then $p_o=0$ and it will not be necessary to calculate $best(v_o)$.

■

So far we showed how the best partial policy is calculated. Will that be the policy of our algorithm? The answer is *No* because we want to allow for some tolerance.

If two policies differ only slightly in the first coordinates of their expected grades, then a minor increase of $h$ is very likely to reverse the order of preferences. Therefore, for a certain policy to be preferred, it should be substantially better (i.e. the difference at some of the coordinates should be greater than $\varepsilon$).

We will define the best partial policy with tolerance $\varepsilon$ and that will be the policy of our algorithm.

## 9. The tolerance ε

We will modify the above algorithm by changing the *best* function. While the initial *best* function returns the best grade, the modified function will return the set of best grades with tolerance ε.

How shall we modify the search for the maximum grade to a search for a set of grades? The previous search looked at the first coordinate and picked the grades with the highest value at that coordinate. The search then went on only within these grades to find the ones with the highest value of the second coordinate and so on until it settles for a single grade. The modified search will pick i) the grades with the highest value of the first coordinate and ii) the grades which are at distance $\varepsilon$ from the maximum value. Let $E_0$ be the initial set of grades. Let in $E_0$ there be $n$ grades, all of them with length $m+1$. We will construct the sequence of grade sets $E_0, \dots, E_{m+1}$ ($E_{i+1} \subseteq E_i$) and the last set $E_{m+1}$ will be the target set of best grades with tolerance $\varepsilon$. Let $E_0=\{G_1, \dots, G_n\}$ and $G_j=g_{j0}, \dots, g_{jm}$. We will also construct the target grade $\alpha$ ($\alpha=\alpha_0, \dots, \alpha_m$). The target set of grades $E_{m+1}$ will be comprised of the grades at distance $\varepsilon$ from $\alpha$.

**Definition 24:** The target grade $\alpha$ and the target set $E_{m+1}$ are obtained as follows:

$$\alpha_0 = \max_{G_j \in E_0} g_{j0}$$

$$E_1=\{ G_j \in E_0 \,/\, \alpha_0 - g_{j0} < \varepsilon \}$$

$$\alpha_1 = \max_{G_j \in E_1} g_{j1}$$

$$E_2=\{ G_j \in E_1 \,/\, (\alpha_0 - g_{j0}) + \gamma.(\alpha_1 - g_{j1}) < \varepsilon \}$$

Here $\gamma$ is again a discount factor. Thus, we have modified the way in which the maximum is calculated. We also need to modify the sum of the grades.

Now the individual grades will be replaced with sets of grades. We will develop all possible combinations and calculate the sum for each combination. The resulting set will be the set of all sums for all possible combinations.

The only remaining thing to do now is to select the next move. We will take the sets of grades provided by the *best* function for the direct successors of $v_t$. Then we will make the union of these sets and from that union we will calculate the set of best grades with tolerance $\varepsilon$. Finally, we will select one of the actions which take us to one of these best grades.

## 10. Is this AI?

Does the algorithm described above satisfy our AI definition? Before that we must say that the algorithm depends on the parameters $h$ and $\varepsilon$. In order to reduce the number of parameters, we will assume that $\varepsilon$ is a function of $h$. For example, this function can be $\varepsilon=h^{-0.5}$.

**Statement 1:** When the value of $h$ is sufficiently high, the described algorithm is sufficiently proximal to the best performing policy.

Let the best performing policy be $P_{best}$, and the policy calculated by the above algorithm with parameter $h$ be $P_h$. Then statement 1 can be expressed as follows:

$$\forall\varepsilon>0\ \exists n\ \ \forall h>n\ (\ |P_{best} - P_h|<\varepsilon\ )$$

Although we cannot prove this statement, we can assume that when $h$ tends to infinity then $P_h$ tends to the best performing policy for the worlds the complexity of which is $k$. When $t$ tends to infinity, $k$ will reach the complexity of the world or tend to infinity. These reflections make us believe that the above statement is true.

## 11. A world with randomness

The first language for description of worlds which discussed here describes deterministic worlds. But, if the world involves some randomness, then the description obtained by using that language would be very inaccurate. Accordingly, we will add randomness to the language for description of worlds. This would improve the language and make it much more expressive.

The new language will also describe the world by a computable function. However, this function will have one additional argument – randomness. By randomness we will mean the result from rolling a dice. Let the complexity of the world be $k$. Then the dice will have $k$ faces and can accordingly return $k$ possible results. The probabilities of occurrence of one of these results will be $p_1,\ \dots\ ,\ p_k$.

**Definition 25:** A model of life until moment $t$ with complexity $k$ will be a world with complexity $k$ and randomness with a length of $t$. We want that life to be generated by that model and that randomness. The randomness will be some word $R$ of length $t$. The $R$ letters will be those from the Turing machine alphabet except $\lambda_0$.

The weight of the model is the probability of occurrence of $R$.

**Definition 26:** The weight of the model will be $p_1^{L_{\lambda_1}(R)}.\ \dots\ .p_k^{L_{\lambda_k}(R)}$.

We will set the probabilities $p_1,\ \dots\ ,\ p_k$ of the model such that the probability of occurrence of $R$ becomes maximal:

$$p_i = \frac{L_{\lambda_i}(R)}{|R|}$$

Thus, we will end up with some low-weight models where the probability of occurrence of the life represented by the model is very low, and some heavy-weight models in which the probability of occurrence is higher.

## 12. A definition with randomness

Similar to the process described above, we will define the best performing policy for the models the complexity of which is $k$. (An important element here is that these models have different weights.) We will develop the policy which represents the limit when $k$ tends to infinity, and that will be the best performing policy. Again, AI will be defined as a computable policy which is sufficiently proximal to the best performing policy.

**Statement 2:** The two AI definitions are identical.

This means that the best performing policy for worlds without randomness is the same as the best performing policy for worlds with randomness. Before we can prove this statement, we need to prove that:

**Statement 3:** If we have some word $\omega$ over the alphabet $\{0, 1\}$ such that the instances of 1 occur with a probability of $p$, and if we make a natural extension of this word, then the next letter will be 1 with probability $p$.

What is a natural extension? Let us take the first (simplest) Turing machine which generates $\omega$. The natural extension will be the extension generated by that Turing machine.

While we cannot prove statement 3, we can offer two ideas about how to prove it:

The first idea is a practical experiment. We will write a program which finds the natural extension of a sequence and then we will run a series of experiments. We will keep feeding into the program various $\omega$ words where 1 occurs with probability $p$. Then we will check the extensions and will calculate the average probability for all these experiments. If the experiments are many and if the average probability obtained from these experiments is $p$, then we can assume that statement 3 is true.

The second idea is to prove the statement by theoretical reasoning. Let us have a computable function $f$ from $\mathbb{N}$ to $\mathbb{N}$. Suppose we start from the number $n$. The resultant sequence will be $\{f^i(n)\}$. We will convert this sequence into sequence $\{b_i\}$ which is made of instances of 0 and 1. The number $b_i$ will be zero iff $f^i(n)$ is an even number. Let $\omega$ be some beginning of $\{b_i\}$. What do we expect the next member of $\{b_i\}$ to be?

Case 1. Sequence $\{b_i\}$ is cyclic and has the form $\omega_1\omega_2^*$. Let $\omega$ be longer than $\omega_1$. Then there is some beginning of $\omega_2$ which is part of $\omega$ and for that beginning the instances of 1 occur with probability $p$.
Case 2. Sequence $\{f^i(n)\}$ has a long beginning in which odd numbers occur with probability $p$. We do not have a reason to expect that the $p$ probability will change.

## 13. A program with randomness

We will develop a program which satisfies an AI definition based on models with randomness. We will proceed in the similar way as above, but with some differences.

We will not search for the first $k$ for which there is a model until moment $t$ with complexity $k$ since such a model exists for very low value of $k$. Instead, we will assume that $k$ is fixed and $k$ is parameter of the algorithm.

The first step will be to find all models of complexity $k$ of vertex $v_t$. The second step will be to run at depth level $h$ across a partial subtree of vertex $v_t$ over i) all discovered models, ii) over all possible actions of the agent and iii) over all probabilities $R_1R_2$, where $R_1$ is the probability of the model and $R_2$ is the probability after $t$. Here $R_1$ is fixed (it is determined by the model), and $R_2$ runs over all possibilities.

The next statement will be similar to statement 1:

**Statement 4:** When the values of $k$ and $h$ are sufficiently high, the described algorithm is sufficiently proximal to the best performing policy.

We assert that when the values of the parameters are sufficiently high, both algorithms will calculate approximately the same policy. However, are the two algorithms equally efficient?

In practice both algorithms are infinitely inefficient, however, the second algorithm is far more efficient than the first one. We will look at three cases:

1. Let us have a simple deterministic world. By *simple* we mean that its complexity $k$ is very low. In this case the first algorithm will be slightly more efficient because it will find the model quickly. The second algorithm will find the same model because the deterministic models are a subset of the non-deterministic ones.

2. Let us have a deterministic world which is not simple, i.e. its complexity $k$ is high. In this case the first algorithm will need a huge amount of time in order to find a model of the world. Moreover, rather than the real model of the world, it will probably find some simplified explanation. That simplified explanation will model the life until moment $t$, but after a few more steps the model will err. The second algorithm will also find a simplified explanation of the world, but that simplified explanation will be non-deterministic. While both algorithms will predict the future with some degree of error, the description which includes randomness will be better and more accurate. Moreover, the description with randomness will be much simpler (with smaller $k$).

3. Let us have a world with randomness. In this case the second algorithm has a major advantage. It will find the non-deterministic model of the world and will begin predicting the future in the best possible way. It may appear that the first algorithm will not get there at all, but this is not the case. It will get there, too, but much later and not so successfully. The non-deterministic model consists of a computable function $f$ and randomness $R$. There exists a computable function $g$ which generates $R$. The composition of $f$ and $g$ will be a deterministic model of the world at moment $t$. Certainly, after a few more steps $g$ will diverge from the actual randomness and $f^o g$ will not be a model of the world anymore. Then we will have to find another function $g$. All this means that a deterministic function can describe a world with randomness, but such description will be very ungainly and will work only until some moment $t$. The non-deterministic model gives us a description which works for any $t$.

The conclusion is that the choice of language for description of the world is very important. Although these two languages provide identical AI definitions, the programs developed on the basis of each language differ substantially in terms of efficiency.

## 14. A world with many agents

The world with randomness can be imagined as a world with one additional agent who plays randomly. Let us assume that there are many agents in the world and each of these agents belongs to one of the following three types:

1. Friends, i.e. agents who help us.
2. Foes, i.e. agents who try to disrupt us.
3. Agents who play randomly.

Let the number of additional agents be $a$ (all excluding the protagonist). Let each additional agent have $k$ possible moves ($k$ is the complexity of the world). We will assume that the protagonist (that's us) will play first and the other agents will play after us in a fixed order. We assume that each additional agent can see everything (the internal state of the world, the model including the number of agents and the type of each agent, i.e. friend or foe, as well as the moves of the agents who have played before him). We will also assume that the agents are very smart and capable to calculate which move is the best and which move is the worst.

The model of the world will again be a Turning machine, but that machine will have more arguments (the internal state of the world and the move of the protagonist, plus the moves of all other agents). The model will also include the type of each agent, i.e. friend or foe. Furthermore, the model of life until moment $t$ will include the moves of all $a$ agents at all steps until $t$.

Once again, we will develop an AI definition on the basis of this new and more complicated language. We will continue with the assumption that the third definition is identical to the previous two. We will also develop a program which looks for a model of the world in the set of worlds with many agents. In the end of the day we will see that the new language is far more expressive: If we have at least one foe in the world this way of describing the world is much more adequate and, accordingly, the AI program developed on the basis of that language is far more efficient.

## 15. Conclusion

We examined three languages for description of the world. On the basis of each language, we developed an AI definition and assumed that all three definitions are the same. Now we will make an even stronger assertion:

**Statement 5:** The AI definition does not depend on the language for description of the world on the basis of which the definition has been developed.

We cannot prove this statement although we suppose that it is true. We also suppose that the statement cannot be proven (similar to the thesis of Church).

Although we assumed that the AI definition does not depend on the language for description of the world, we kept assuming that the program which satisfies this definition strongly depends on the choice of language. The comparison between the first two languages clearly demonstrated that the second language is far more expressive and produces a far more efficient AI.

Let us look at one more language for description of worlds – the language described in Dobrev (2022, 2023). That language describes the world in a far more efficient way by defining the term

'algorithm'. The term 'algorithm' enables us plan the future. For example, let us take the following: 'I will wait for the bus until it comes. Then I will go to work and will stay there until the end of the working hours.' These two sentences describe the future through the execution of algorithms. If we are to predict the future only by running $h$ possible steps, then $h$ will necessarily become unacceptably large.

The language described in Dobrev (2022, 2023) is far more expressive and lets us hope that it can be used to produce a program which satisfies the AI definition and which is efficient enough to work in real time.

## 16. Acknowledgements

I wish to thank my colleague Ivan Soskov (Soskov 2013), with whom we discussed the efficiency of AI several years ago. Let me also thank my colleagues Dimitar Dimitrov (Dimitrov 2025) and Joan Karadimov (Karadimov 2025). This paper was inspired in a conversation I had with them. Now I wish to apologize to Marcus Hutter and Shane Legg for the unfair allegation in my paper Dobrev (2013) that they had used some work of mine without citing it. Later on, when I carefully read Marcus Hutter's paper Hutter (2000) I realized that he had published there the main concepts well before I did.